# Fast Classification of Large Time Series Datasets

Muhammad Marwan Muhammad Fuad

Coventry University
Coventry CV1 5FB, UK
ad0263@coventry.ac.uk

**Abstract:** Time series classification (TSC) is the most import task in time series mining as it has several applications in medicine, meteorology, finance cyber security, and many others. With the ever increasing size of time series datasets, several traditional TSC methods are no longer efficient enough to perform this task on such very large datasets. Yet, most recent papers on TSC focus mainly on accuracy by using methods that apply deep learning, for instance, which require extensive computational resources that cannot be applied efficiently to very large datasets. The method we introduce in this paper focuses on these very large time series datasets with the main objective being efficiency. We achieve this through a simplified representation of the time series. This in turn is enhanced by a distance measure that considers only some of the values of the represented time series. The result of this combination is a very efficient representation method for TSC. This has been tested experimentally against another time series method that is particularly popular for its efficiency. The experiments show that our method is not only 4 times faster, on average, but it is also superior in terms of classification accuracy, as it gives better results on 24 out of the 29 tested time series datasets. .

## 1  Introduction

A time series $T = (\langle t_1, v_1 \rangle, \langle t_2, v_2 \rangle, \ldots, \langle t_n, v_n \rangle)$ is an ordered collection of observations measured at timestamps $t_n$. If the values $v$ are single elements, usually real numbers, the time series is called *univariate*, if they are vectors, the time series is called *multivariate*

Time series data appear in a wide variety of applications, such as medicine, finance, meteorology, and many others. The recent proliferation of internet of things (IoT) with sensors capturing very large amount of data, in the form of time series, added another rich source of time series data

Time series mining addresses several tasks such as clustering, segmentation, query-by-content, anomaly detection, prediction, and others.

The most important time series mining task is classification. Time series classification (TSC) has a broad variety of applications in medicine, industry, meteorology, finance, and many other domains. This variety of applications is the reason why TSC has gained increasing attention over the last two decades [2] [5] [13] [12] [19].

Although there are several general purpose classification methods in the data mining and machine learning literature, time series mining has its own classification methods. This is because of the ordered nature of the time series, in addition to the high correlations that exist among the values.

One of the difficulties in handling time series data is their high dimensionality. One approach that has been widely used in time series mining to overcome the problem of high dimensionality, also known as *curse of dimensionality*, is to use time series representation methods, and then to apply the different tasks on these representations of time series, instead of applying them directly to raw data.

There has been a plethora of time series representation methods in the literature, to mention a few; the Piecewise Aggregate Approximation (PAA) [9] [20], Adaptive Piecewise Constant Approximation (APCA) [14], and the Symbolic Aggregate Approximation – SAX [16] [17]

Most of these methods include dimensionality reduction; i.e. they transform the raw time series from their original space to one with a lower dimensionality, and then the different time series mining tasks can be performed in these lower dimensional spaces under certain conditions.

Although these representation methods can have a better performance in TSC than when applied to raw data directly, their application, for a variety of reasons, is not suitable for very large time series datasets. There are however, some time series representation methods which made efficiency its main objective. This made these methods easier to apply to very large time series datasets.

In this paper we present a new time series representation method that is designed particularly for efficiently classifying very large time series datasets. We will show experimentally how this new method is not only quite efficient, but also gives good results, in terms of TSC accuracy. The secret behind this performance is its simplicity and the way it calculates the distance between two time series, as our method only considers certain points and discards the others.

Technically, our method is not a dimensionality reduction method, as the dimensionality of the represented time series data is almost the same as that of the original time series. This may seem to put our method at disadvantage compared with other time series representation methods, but we will show later how our method allows compressing the data, which substantially reduces storage requirements, even much more than traditional time series representation methods

The rest of the paper is organized as follows: in Section 2 we present related background. The new method is presented in Section 3 and validated experimentally in Section 4. We conclude with Section 5.

## 2 Background

Time series data are ubiquitous. Time series mining has several applications in medicine, finance, meteorology, astronomy, and many others. Time series manning handles several tasks such as clustering, motif discovery, rule discovery, query-by-content, and anomaly detection. However, the most important task is classification.

Given a time series dataset $D$ of $m$ time series, each of $n$ dimensions. Each time series $T_i$, $i \in \{1,2,...,m\}$ is associated with a class label $L(T_i)$; $L(T_i) \in \{1,2,...,c\}$. For a new set $U$ of unlabeled time series, the purpose of time series classification is to map each time series in $U$ to one of the classes in $\{1,2,...,c\}$. This is done based on the knowledge the classification algorithm learned from $D$ during a training stage, as is the case with other classification algorithms in data mining

As with other time series mining tasks, TSC takes advantages of the particularity of time series data, which is their natural temporal ordering of the attributes [1], in addition to their high correlation

The most common TSC method is the *k-nearest-neighbor* (kNN). The intuition behind this method is that the new object should be classified based on the $k$ objects closer to it, one case of particular interest is when $k$=1. This "closeness" between the new object and each of these $k$ objects is measured using the notion of *distance*. Two very popular examples that are widely used in TSC are the *Dynamic Time Warming* (DTW) and the *Euclidean distance*. There are a few differences between these two measures, which affects their use in TSC. The Euclidean distance is applied to time series of the same lengths, unlike DTW, which can be applied to time series of different lengths too. However, the Euclidean distance has a few inconveniences: it is sensitive to noise and to shifts on the time axis and thus lacks elasticity [3], [11], [15]. DTW is known to give better, or even much better, results than the Euclidean distance in several time series data mining tasks such as classification and clustering. Its applicability to time series of different lengths is another advantage that it has over the Euclidean distance. However, its main drawback is that its complexity $O(n^2)$ is very high compared with that of the Euclidean distance $O(n)$. This makes its use inefficient in large datasets, mainly when the dimensionality of the time series is high. A few researchers proposed some variants of DTW in order to alleviate this effect of high complexity, but even with those variants, this issue remained an obstacle to using DTW. This issue is particularly important, because several time series datasets in use nowadays are very large with hundreds or thousands of dimensions, which makes the use of DTW almost impossible as it would take an extremely long time.

Distance computation is a time consuming process, so one of the objectives of time series mining is to reduce the number of distance calculations for whatever time series mining tasks we are performing. For this reason, reducing the number of distance calculations, or using distance with low complexity, could substantially speed up performing time series mining tasks.

Although TSC can be applied to raw data, it is usually applied to representations of these data. This has several advantages as it is usually more efficient and gives more accurate classification results, since in many cases, these representations of the data have a smoothing effect. Another advantage is that these representations of time series require less storage than storing the original raw data. However, applying TSC to very large datasets is still a challenging task. The main reason for this, in our opinion, is that these representation methods focus on reducing dimensionality, but most of them adopt an expensive to compute distances, which becomes a bottleneck when applied to very large dataset.

We believe this problem should be approached by focusing on simplicity, simplicity in the representation itself, and also in the distance used in the TSC.

As mentioned in the introduction, the method we present in this paper does not reduce the data dimensionality, as the time series representation the date is transformed into has a dimensionality that is very close to that of the original one. What we do however is to discretize each value of the raw data into one of three values. This has two benefits; the first is the ability to take full advantage of data compression to substantially reduce data storage requirements. The second benefit is that this simple representation will allow using an inexpensive distance. The other advantage of our method is that we only calculate what is "worth considering" of the time series and discard the other parts. These two features of our method make it very efficient to perform TSC on very large time series datasets.

## 3 The Fast Discretization (FD) Representation

As with other representation methods, our new method, which we call the *Fast Discretization* method has two components; the first is the representation itself, which is the discretization part, and the second is the distance we use with our representation

**3.1 The FD Representation**

Let $D$ be a dataset of $m$ time series of $n$ dimensions (the length of the time series), i.e. $T_i = (\langle t_1, v_{i,1}\rangle, \langle t_2, v_{i,2}\rangle, \ldots, \langle t_n, v_{i,n}\rangle)$ where $1 \leq i \leq m$, and let $\mu$ be the mean of all the values $v_{m,n}$ and $\sigma$ be their standard deviation, i.e. the mean and standard deviation are calculated over $D$ and not over individual time series. Let $w$ (an integer) be the sliding window size and $\alpha$ (a real number) is a parameter (we will show later in the experimental section how to choose the values of both $w$ and $\alpha$)

The *FD* discretization of a time series $T = (\langle t_1, v_1\rangle, \langle t_2, v_2\rangle, \ldots, \langle t_n, v_n\rangle)$ is performed by sliding a window $w$ from from $t_j = 1$ to $t_j = n - w + 1$. Each segment $s = (\langle t_j, v_j\rangle, \ldots, \langle t_{j+w-1}, v_{j+w-1}\rangle)$ is processed as follows:

$$f(v_j) = \begin{cases} +1 & if\ v_j \geq \mu + \alpha \times \sigma \\ -1 & if\ v_j \leq \mu - \alpha \times \sigma \\ 0 & otherwise \end{cases} \quad (1)$$

Let $s^+$ be the number of $v_j$ in $s$ whose value is +1 according to Eq. (1) and $s^-$ be the number of $v_j$ in $s$ whose value is -1 according to Eq. (1). The discretization of the segment $s$ is given as follows:

$$F(s) = \begin{cases} +1 & if\ s^+ > s^- \\ -1 & if\ s^+ < s^- \\ 0 & if\ s^+ = s^- = 0 \\ tb & if\ s^+ = s^- \neq 0 \end{cases} \quad (2)$$

Where the value of $tb$ (tie breaker) is given by the following equation:

$$tb = \begin{cases} +1 & if\ abs\ (v_s^+) > abs\ (v_s^-) \\ -1 & if\ abs\ (v_s^+) < abs\ (v_s^-) \end{cases} \quad (3)$$

Where $v_s^+$ is the largest value of the points in $s$ whose representations are +1 according to Eq.1, and $v_s^-$ is the smallest value of the points in $s$ whose representations are -1 according to Eq. 1,

The $FD$ discretization of a time series $T$ is the concatenation of $F(s)$. Notice that the length of $FD\ (T)$ is $n - w + 1$. Given that $w$ is small (as we will see later), the $FD$ discretization cannot actually be considered a dimensionality reduction method.

Informally, the representation can be explained in simple words as follows, a sliding window scans the original raw data, the window moves one timestamp at a time, all the data points within the current window whose values are higher than a certain threshold, are represented as +1, and all those whose values are lower than the opposite of this threshold are represented as -1. The representation of the segment itself is determined using a vote of the number of points that were represented as +1 and those which were represented as -1. Ties are broken by comparing the maximum absolute value of the points which were represented as +1 and those which were represented as -1.

**3.2 The Fast Distance (FDist)**

Given two time series $T = (\langle t_1, v_1 \rangle, \langle t_2, v_2 \rangle, \dots, \langle t_n, v_n \rangle)$ and $R = (\langle t_1, r_1 \rangle, \langle t_2, r_2 \rangle, \dots, \langle t_n, r_n \rangle)$, whose $FD$ representations are :

$FD(T) = [\bar{v}_1, \bar{v}_2, \dots, \bar{v}_{n-w+1}]$, $FD(R) = [\bar{r}_1, \bar{r}_2, \dots, \bar{r}_{n-w+1}]$, respectively The $FDist$ between $FD(T)$ and $FD(R)$ is given by:

$$FDist\big(FD(T), FD(R)\big) = 1 - \frac{sim\big(FD(T), FD(R)\big)}{n - w + 1} \quad (4)$$

Where

$$sim(FD(T), FD(R)) = \sum_{i=1}^{n-w+1} I(\bar{\bar{v}}_i = \bar{\bar{r}}_i)$$

Here, $I(.)$ denotes the indicator function, which is equal to 1 if the argument is true, and, 0 otherwise, and where $\bar{\bar{v}}_i$ ($\bar{\bar{r}}_i$) are the values $\bar{v}_i(\bar{r}_i)$ that are either equal to 1 or -1

**3.3 Illustrating Example**

We use the following time series to show how our method is applied. The time series we use in this example is the first time series from dataset "FacesUCR". This is one of the datasets we use in the experimental section of this paper. For this dataset "FacesUCR", $\mu = -6.5700E - 17$ and $\sigma = 0.9962$. We use $\alpha = 0.1$ (this will be discussed further in the experimental section). So for this dataset, Eq. 1 is written as:

$$f(v_j) = \begin{cases} +1 & if\ v_j \geq 0.0996 \\ -1 & if\ v_j \leq -0.0996 \\ 0 & otherwise \end{cases} \quad (5)$$

The following is the first 32 values of this dataset. The values are rounded to save space:

$T = [0.48, 0.48, 0.48, 0.45, 0.35, 0.21, 0.10, 0.06, 0.02, -0.03, -0.07, -0.02, 0.04, 0.06,$
$0.05, 0.01, -0.07, -0.13, -0.14, -0.15, -0.26, -0.37, -0.35, -0.31, -0.24, -0.17,$
$-0.09, 0.00, 0.11, 0.30, 0.49, 0.58]$

Sliding a window $w = 4$ over $T$ will generate $n - w + 1 = 29$ segments: $s_1 = [0.48, 0.48, 0.48, 0.45]$, $s_2 = [0.48, 0.48, 0.45, 0.35]$, ..., $s_{29} = [0.11, 0.30, 0.49, 0.58]$.

In the following we will show some examples of discretizing some of these segments using our method.

In order to discretize $s_1$ we apply Eq. 5, to the values of this segment, we get the following: $[+1, +1, +1, +1]$. We thus have $s_1^+ = 4$, $s_1^- = 0$, so $F(s_1) = +1$

$s_8 = [0.06, 0.02, -0.03, -0.07]$. By applying Eq.5, the values of this segment are discretized as $[0, 0, 0, 0]$, so $F(s_8) = 0$

$s_{26} = [-0.17, -0.09, 0.00, 0.11]$. The values are discretized as $[-1, 0, 0, +1]$. We notice in this case that $s_1^+ = 1$, $s_1^- = 1$, so we have a tie. We apply Eq.3 to break the tie: $abs\ (0.11) < abs\ (-0.17)$, so $F(s_{26}) = -1$

The above examples cover all different scenarios, so the other segments are discretized in a similar manner to get the $FD$ representation of this time series:

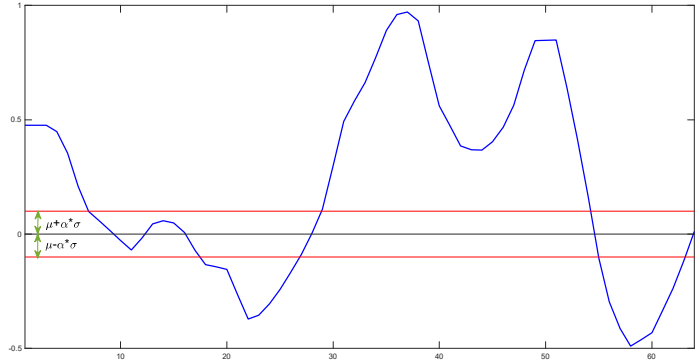

**Fig. 1.** Time series $T$ given in the illustrated example.

$$FD(T) = [1,1,1,1,1,1,1,0,0,0,0,0,0,0,-1,-1,-1,-1,-1,-1,-1,-1,-1,1,-1,-1,-1,1,1]$$

To complete our example, say we have another time series $S$ whose $FD$ representation is:

$$FD(S) = [-1,1,1,1,-1,1,0,0,0,0,0,0,0,-1,0,-1,1,0,-1,0,1,-1,1,-1,0,0,0,1,-1]$$

Then:

$$sim(FD(T), FD(S)) = \sum_{i=1}^{n-w+1} I(\bar{\bar{v}}_i = \bar{\bar{r}}_i) = \frac{8}{29}$$

$$FDist(FD(T), FD(R)) = 1 - \frac{8}{29} = 0.7241$$

Notice that when we computed $sim(FD(T), FD(S))$ we only considered the matches where the two series had a value of either 1 or -1. This happens when $i = 2, 3, 4, 6, 16, 19, 22, 28$

Time series $T$ is shown in Fig1.

### 3.4 Remarks

    i-    Notice that a segment $s$ is represented as 0 if none of its points are represented as +1 or -1. The underlying idea in this case is that "there is no information in this segment that is worth considering" when

ii- calculating the distance between two time series represented by our method. This is one of the principles of our method.

ii- Since the sliding window is moving one position at a time, the data points are actually over-represented. This is a characteristic of our representation that we take into account when calculating the distance between two time series represented by our method.

iii- Using a sliding window for representation has its advantages compared with segmentation approaches methods which represent each segment independently from the other, with all the adverse consequences of such methods, the most important of which is that the transition between one segment and another is not represented when segmentation is used. In addition to other issues related to segment length.

iv- Our representation, and this is an advantage, is insensitive to local fluctuations. In other words, it does not differentiate between having one local extreme point and several neighboring extreme points. This is reflected by the fact that having one point only with a value +1(-1) in a segment is the same as having several points with a value +1(-1). It is also torrent to small fluctuations in local extreme points, as can be illustrated in Fig. 2, which shows four time series of length 32. The one on the top is $T$ that we defined at the beginning of this section. Each of the other three time series below it are ones where the value of one or two local extreme points was changed compared with $T$ (indicated by the rectangles). The $FD$ representation of the four time series is the same which is:

$$[1,1,1,1,1,1,0,0,0,0,0,0,0,-1,-1,-1,-1,-1,-1,-1,-1,-1,1,-1,-1,-1,1,1]$$

This characteristic of our method makes it less sensitive to local fluctuations, unlike the Euclidean distance, for instance.

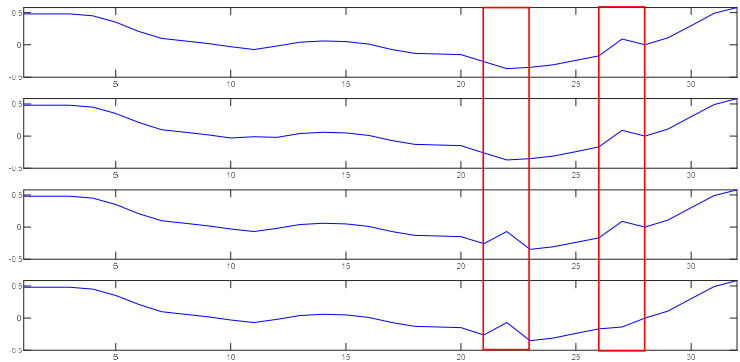

**Fig. 2.** The four time series described in Section 3.4 Remarks (iv). They have different local extreme points, but their $FD$ representation is that same

### 3.5 Complexity Considerations

Although, as we indicated earlier, our method is not technically a dimensionality reduction method, because the dimensionality of the $FD$ representation is almost the same as that of the original time series, the actual storage requirement is much less because our representation uses three integers only 1, 0, and -1. These require 2 bits each (two bits can encode 4 values), compared with real numbers used to store raw data, which require more than that on any computer system. In addition to that, because change in time series is usually not sudden, the common lossless compression technique *Run Length Encoding* (RLE) [8] can be effectively used to substantially reduce storage requirements. For instance, the $FD$ representation of time series $T$ in the illustrating example in Section 3.3, which is
$[1,1,1,1,1,1,1,0,0,0,0,0,0,0,-1,-1,-1,-1,-1,-1,-1,-1,-1,1,-1,-1,-1,1,1,1]$ can be compressed as **7**#1, **7**#0, **9**#−1, **1**#1, **3**#−1, **2**#1, which requires much less storage than the original time series.

The computational complexity when computing $FDist(.)$ (Eq. (4)), which boils down to computing $sim(.)$ is very low, since it is completely based on comparing values, whose computational complexity is very low.

## 4  Experiments

We tested our method in a 1NN classification task, which we presented in Section2, using all the large datasets that we found in *UCR Time Series Classification Archive* [4]. This archive is the most widely used in the literature. This archive contains 128 time series datasets of different sizes (number of instances in the dataset) and dimensions (the length of the time series in the datasets).

Determining what is considered "large" is not straightforward. However, in order to get a reasonable number of dataset to test our method on, we considered that any dataset with more than 1000 instances to be large. Of the 128 datasets in the archive, 28 fulfil this stipulation (there were two other datasets but we excluded them because they have missing values). We added another one, whose number of instances is very close to 1000 (995), so the final number is 29. This is a reasonable number to test on, especially that our method is not a general purpose one, but it targets large datasets explicitly. These 29 datasets are shown in Table 1. As we can see from the table, these datasets are generated from different sources. They also have a variety of classes and lengths. Their sizes, vary between 995 up to 16800 instances.

In order to test the performance of our method, we compared it to another, very popular, method of TSC, which is the *Symbolic Aggregate approXimation* (SAX) [16] [17]. SAX is one of the most powerful symbolic representation methods of time series. However, the reason why we have chosen it is not its popularity, but because SAX is very efficient, which is the reason why it is so popular. In fact a variation of it, *i*SAX [22], is used to index and mine very large time series datasets, so it is in the same line with our method.

SAX is based on another method; the *Piecewise Aggregate Approximation* - PAA [9] [20] and it is applied as follow:

**Table 1.** The datasets used in the experiments, their type, their size (number of time series in each dataset), the number of classes, and the length of the time series

| Dataset | Type | Size | Classes | Length |
|---|---|---|---|---|
| ChlorineConcentration | Sensor | 3840 | 3 | 166 |
| CinCECGTorso | Sensor | 1380 | 4 | 1639 |
| ECG5000 | ECG | 4500 | 5 | 140 |
| ElectricDevices | Device | 7711 | 7 | 96 |
| FaceAll | Image | 1690 | 14 | 131 |
| FacesUCR | Image | 2050 | 14 | 131 |
| FordA | Sensor | 1320 | 2 | 500 |
| InsectWingbeatSound | Sensor | 1980 | 11 | 256 |
| ItalyPowerDemand | Sensor | 1029 | 2 | 24 |
| Mallat | Simulated | 2345 | 8 | 1024 |
| MoteStrain | Sensor | 1252 | 2 | 84 |
| NonInvasiveFetalECGThorax1 | ECG | 1965 | 42 | 750 |
| NonInvasiveFetalECGThorax2 | ECG | 1965 | 42 | 750 |
| Phoneme | Sensor | 1896 | 39 | 1024 |
| StarLightCurves | Sensor | 8236 | 3 | 1024 |
| Symbols | Image | 995 | 6 | 398 |
| TwoLeadECG | ECG | 1139 | 2 | 82 |
| TwoPatterns | Simulated | 4000 | 4 | 128 |
| UWaveGestureLibraryAll | Motion | 3582 | 8 | 945 |
| UWaveGestureLibraryX | Motion | 3582 | 8 | 315 |
| UWaveGestureLibraryY | Motion | 3582 | 8 | 315 |
| UWaveGestureLibraryZ | Motion | 3582 | 8 | 315 |
| Wafer | Sensor | 6164 | 2 | 152 |
| Yoga | Image | 3000 | 2 | 426 |
| Crop | Image | 16800 | 24 | 46 |
| FreezerRegularTrain | Sensor | 2850 | 2 | 301 |
| FreezerSmallTrain | Sensor | 2850 | 2 | 301 |
| MixedShapesRegularTrain | Image | 2425 | 5 | 1024 |
| MixedShapesSmallTrain | Image | 2425 | 5 | 1024 |

1- The time series are normalized.
2- The dimensionality of the time series is reduced by using PAA
3- The PAA representation of the time series is discretized.
4- The last step of SAX is using the following distance:

$$MINDIST(\hat{S}, \hat{T}) = \sqrt{\frac{n}{r}} \sqrt{\sum_{i=1}^{r}(dist(\hat{s}_i, \hat{t}_i))^2} \quad (6)$$

Where $n$ is the length of the original time series, $r$ is the number of segments, $\hat{S}$ and $\hat{T}$ are the symbolic representations of the two time series $S$ and $T$, respectively, and where the function $dist(\ )$ is implemented by using the appropriate lookup table.

$MINDIST$ is the secret behind the efficiency of SAX, because it applies pre-computed lookup tables to find the values $dist(\hat{s}_i, \hat{t}_i)$. This makes $MINDIST$ easier to compute than most other distances for time series, which need to do calculations on real numbers. However, our distance, $FDist$ is yet significantly faster than $MINDIST$ for several reasons; it still avoids computations, just like $MINDIST$, because of the way time series are discretized in our method, which does not use real numbers. But $FDist$ also has other advantages that $MINDIST$ does not have. It does not include any operation like raising to the power of 2 or calculating the square root that $MINDIST$ applies. These are expensive operations - their latency time, measured by the number of cycles the processor takes to perform different arithmetic operations, the former requires a latency time of 24 cycles and the latter requires 209 cycles [23]. Besides, $FDist$ only calculates "important" values and discards the others completely, unlike $MINDIST$.

SAX has one parameter, which is the alphabet size. However, the authors of SAX do not train their method to find the best value of alphabet size for each dataset. They say

**Table 2.** The average time, in seconds, of 10 runs of the classification task on the 29 datasets presented in Table 1, using our method and SAX. The dataset for which the average run time of one method is shorter is shown in bold

| Dataset | FD | SAX | Speed Gain |
|---|---|---|---|
| ChlorineConcentration | **4.49** | 8.14 | 1.8129 |
| CinCECGTorso | **2.54** | 35.33 | 13.9094 |
| ECG5000 | **4.95** | 21.37 | 4.3172 |
| ElectricDevices | **144.37** | 329.55 | 2.2827 |
| FaceAll | **2.43** | 4.02 | 1.6543 |
| FacesUCR | **0.12** | 0.17 | 1.4167 |
| FordA | **3.80** | 15.01 | 3.9500 |
| InsectWingbeatSound | **1.71** | 6.54 | 3.8246 |
| ItalyPowerDemand | **0.13** | 0.40 | 3.0769 |
| Mallat | **0.33** | 1.42 | 4.3030 |
| MoteStrain | **0.16** | 0.24 | 1.5000 |
| NonInvasiveFetalECGThorax1 | **24.40** | 165.17 | 6.7693 |
| NonInvasiveFetalECGThorax2 | **24.48** | 146.56 | 5.9869 |
| Phoneme | **5.32** | 40.74 | 7.6579 |
| StarLightCurves | **99.39** | 663.06 | 6.6713 |
| Symbols | 0.47 | **0.42** | 0.8936 |
| TwoLeadECG | 0.15 | **0.13** | 0.8667 |
| TwoPatterns | **9.72** | 19.24 | 1.9794 |
| UWaveGestureLibraryAll | **37.81** | 205.49 | 5.4348 |
| UWaveGestureLibraryX | **10.94** | 36.17 | 3.3062 |
| UWaveGestureLibraryY | **10.75** | 36.38 | 3.3842 |
| UWaveGestureLibraryZ | **13.06** | 61.20 | 4.6861 |
| Wafer | **13.54** | 33.08 | 2.4431 |
| Yoga | **4.39** | 24.47 | 5.5740 |
| Crop | **214.48** | 459.17 | 2.1409 |
| FreezerRegularTrain | **2.09** | 7.67 | 3.6699 |
| FreezerSmallTrain | **1.04** | 1.15 | 1.1058 |
| MixedShapesRegularTrain | **11.52** | 87.38 | 7.5851 |
| MixedShapesSmallTrain | **3.86** | 19.57 | 5.0699 |

in [18] that any value of 3 or 4 works well for most time series datasets so they choose the alphabet size to be 4 for all datasets for simplicity. When we applied SAX in the experiments, we chose the same value for the alphabet size recommended by its authors

Our method has one parameter too, which is $\alpha$. We also adopt a similar approach and we use the same value of $\alpha = 0.01$ for all datasets. We do this for simplicity, but also to make an unbiased comparison with SAX.

Comparing the run times of the two methods will be done in a traditional way using wall clock time. To avoid fluctuations in run time that usually appear due to different factors, we run each method, on each dataset, 10 times and we calculate the average run time of these 10 runs.

The experiments were run on Windows 10 with an Intel Core i7-6600U CPU @ 2.60GHz and 16 GB of RAM

As we can see from Table 2, FD is faster than SAX on almost all datasets. On average it is 4 times faster. Given the different features these datasets have, it is not easy to determine why FD is faster on some datasets more than on others, although we

**Table 3.** The classification errors of FD and SAX on the datasets presented in Table 1

| Dataset | SAX | FD |
| --- | --- | --- |
| ChlorineConcentration | 0.742 | **0.482** |
| CinCECGTorso | 0.304 | **0.277** |
| ECG5000 | 0.195 | **0.097** |
| ElectricDevices | 0.878 | **0.545** |
| FaceAll | 0.571 | **0.305** |
| FacesUCR | 0.476 | **0.251** |
| FordA | 0.378 | **0.348** |
| InsectWingbeatSound | 0.494 | **0.449** |
| ItalyPowerDemand | 0.459 | **0.373** |
| Mallat | 0.668 | **0.311** |
| MoteStrain | 0.277 | **0.215** |
| NonInvasiveFetalECGThorax1 | 0.908 | **0.414** |
| NonInvasiveFetalECGThorax2 | 0.871 | **0.370** |
| Phoneme | 0.946 | **0.929** |
| StarLightCurves | 0.165 | **0.146** |
| Symbols | 0.354 | **0.220** |
| TwoLeadECG | 0.475 | **0.310** |
| TwoPatterns | **0.307** | 0.546 |
| UWaveGestureLibraryAll | **0.062** | 0.076 |
| UWaveGestureLibraryX | 0.414 | **0.387** |
| UWaveGestureLibraryY | **0.463** | 0.474 |
| UWaveGestureLibraryZ | **0.420** | 0.478 |
| Wafer | 0.006 | **0.004** |
| Yoga | 0.390 | **0.231** |
| Crop | 0.918 | **0.595** |
| FreezerRegularTrain | 0.496 | **0.413** |
| FreezerSmallTrain | 0.330 | **0.327** |
| MixedShapesRegularTrain | 0.231 | **0.209** |
| MixedShapesSmallTrain | **0.233** | 0.254 |
| | 5/29 | **24**/29 |

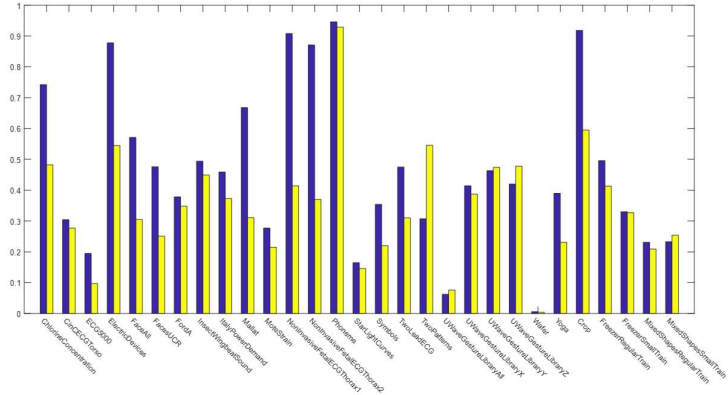

**Fig. 3.** Bar chart comparisons of the classification errors between FD (in yellow) and SAX (in blue) on the 29 tested datasets. The figure shows the superior performance of FD over SAX.

notice that the fastest gain was with CinCECGTorso , which happens to be the dataset with the longest time series, as we see from Table 1, but this is not enough evidence of correlation

In addition to comparing the efficiency of the two methods by measuring the run times of each of them, which is the main motivation for developing our method, we also need to compare the classification accuracy of the two methods on each dataset to see if any improvement in efficiency comes at the expense of degrading classification accuracy. The results of the experiments are shown in Table 3 and Fig. 3

As we can see from Table 3 and Fig.3, the classification accuracy of FD is much better than that of SAX, as FD gives a smaller classification error in 24 out of the 29 datasets. For several datasets, the difference in the classification error is substantial

Finally, Fig. 4 summarizes our previous findings as it shows, on one plot, a comparison of the classification errors for FD versus SAX on all 29 datasets tested. The bottom-right region of the figure shows where the classification errors of FD are lower than those of SAX, whereas the top-left region of the figure shows where the classification errors of SAX are lower than those of FD

## 5  Conclusion

In this paper we presented a new time series representation method, FD, which is designed to classify very large time series data very efficiently. The method uses a basic representation of the data. This is further enhanced by applying a simple distance to classify the time series. The results of all that is a very efficient classification method. This was tested experimentally by comparing our method with another time series representation method, SAX, that is highly popular in the literature, mainly because of its efficiency, which is the reason why we tested our method against it. The experiments show that our method is on average 4 times faster than SAX. Moreover, the interesting thing is that our method is not only faster than SAX, but it also outperforms SAX in terms of classification accuracy as it gave a smaller classification error than SAX in 24 out of the 29 tested datasets.

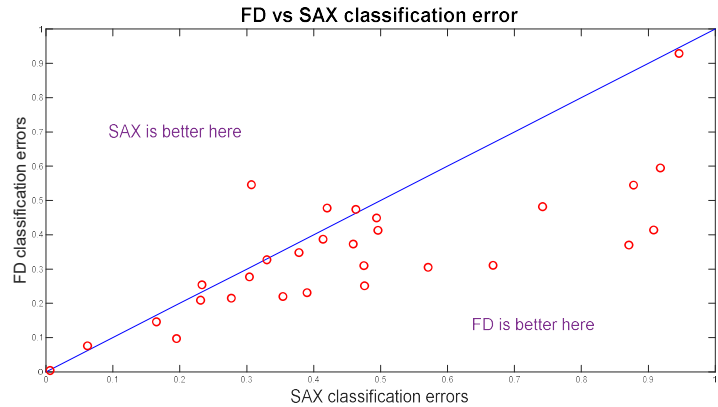

**Fig. 4.** Comparison of the classification errors of FD versus SAX. The bottom-right region is where FD performs better, and the top-left region is where SAX performs better.

It is important to understand, when evaluating our method, or other similar methods such as SAX, that the main objective of these methods is speed. As we can see, performing the classification task does not require any training, and it is performed in seconds. This is different from other TSC methods, mainly those that are based on deep learning [6] [7] [10], which may give a better, or even much better classification accuracy, but they require long training using extensive computational resources. Besides, the purpose of training in such methods is to determine the values of parameters related to a particular dataset, so the results cannot be generalized to new datasets, unlike our method.

Our next direction of future research is to apply our method to other time series mining tasks, mainly clustering, which is also an important task in time series mining and it also requires efficient methods to cluster very large time series data. Query-by-content is also another task that could benefit from our method, whether when applied to the original time series data, or, and we are particularly interested in this, when applied to data that has been transformed into time series, such as genomic data [21]

## References

1. Bagnall, A., Lines, J., Bostrom, A., Large, J., and Keogh, E.: The great time series classification bake off: a review and experimental evaluation of recent algorithmic advances. Data Min Knowl Disc 31, 606–660 (2017)

2. Baydogan, M., Runger, G., Tuv, E.: A bag-of-features framework to classify time series. IEEE Trans Pattern Anal Mach Intell 25(11):2796–2802 (2013)
3. Chu, S., Keogh, E., Hart, D., Pazzani, M.: Iterative deepening dynamic time warping for time series. In Proc 2 nd SIAM International Conference on Data Mining (2002)
4. Dau, H.A., Keogh, E., Kamgar, K., Yeh, C.M., Zhu, Y., Gharghabi, S., Ratanamahatana, C.A., Chen, Y., Hu, B., Begum, N., Bagnall, A., Mueen, A., Batista, G., & Hexagon-ML: The UCR Time Series Classification Archive. URL https://www.cs.ucr.edu/~eamonn/time_series_data_2018/ (2019).
5. Fawaz, H.I., Forestier, G., Weber, J., Idoumghar, L., Muller, P.A.: Adversarial attacks on deep neural networks for time series classification. In Proceedings of the 2019 International Joint Conference on Neural Networks (IJCNN), Budapest, Hungary, 14–19 July (2019)
6. Fawaz H. I, Forestier G., Weber J., Idoumghar L., Muller PA.: Deep learning for time series classification: a review. Data Min Knowl Discov (2019)
7. Fawaz, H.i, Lucas, B., Forestier, G. et al. InceptionTime: Finding AlexNet for time series classification. Data Min Knowl Disc 34, 1936–1962 (2020).
8. Golomb, S.W.:Run-length Encodings. IT(12), pp.399—401, No. 7, July (1966).
9. Hatami, N., Gavet, Y. & Debayle, J.: Bag of recurrence patterns representation for time-series classification. Pattern Anal Applic 22, 877–887 (2019)
10. Hatami N., Gavet Y., Debayle J.: Classification of time-series images using deep convolutional neural networks. In: International conference on machine vision (2017)
11. Kadous, M. W.: Learning Comprehensible Descriptions of Multivariate Time Series. In Proc. of the 16 th International Machine Learning Conference (1999).
12. Karim, F., Majumdar, S., Darabi, H., Chen, S.: LSTM fully convolutional networks for time series classification, IEEE Access 1-7 (2017)
13. Keogh, E., Chakrabarti, K., Pazzani, M. and Mehrotra: Dimensionality reduction for fast similarity search in large time series databases. J. of Know. and Inform. Sys. (2000)
14. Keogh, E., Chakrabarti, K., Pazzani, M., and Mehrotra, S.: Locally adaptive dimensionality reduction for similarity search in large time series databases. SIGMOD pp 151-162 (2001)
15. Kollios, G., Vlachos, M. and Gunopulos, G.: Discovering similar multidimensional trajectories. In Proc 18 th International Conference on Data Engineering (2002).
16. Lin, J., Keogh, E., Lonardi, S., Chiu, B. Y.: A symbolic representation of time series, with implications for streaming algorithms. DMKD 2003: 2-11(2003)
17. Lin, J., Keogh, E., Wei, L., and Lonardi, S.: Experiencing SAX: a novel symbolic representation of time series. Data Min. Knowl. Discov., 15(2), (2007)
18. Lin, J., Khade, R. & Li, Y.: Rotation-invariant similarity in time series using bag-of-patterns representation. J Intell Inf Syst 39, 287–315 (2012)
19. Wang, Z., Yan, W., Oates, T.: Time series classification from scratch with deep neural networks: A strong baseline. In Proc. Int. Joint Conf. Neural Netw. (IJCNN), May 2017, pp. 1578 1585 (2017)
20. Yi, B. K., and Faloutsos, C.: Fast time sequence indexing for arbitrary Lp norms. Proceedings of the 26th International Conference on Very Large Databases, Cairo, Egypt (2000)
21. Shieh, J. and Keogh, E. 2009. Shieh J, Keogh E. $i$SAX: disk-aware mining and indexing of massive time series datasets. Data Min Knowl Discov.;19(1):24–57 (2009)
22. Shieh, J. and Keogh, E. $i$SAX : indexing and mining terabyte sized time series. In Proceeding of the 14th ACM SIGKDD international conference on Knowledge discovery and data mining. ACM, 623–631 (2008).
23. Schulte, M.J. ,Lindberg, M. and Laxminarain, A. :Performance Evaluation of Decimal Floating-point Arithmetic in IBM Austin Center for Advanced Studies Conference, February (2005).